\documentclass[journal]{IEEEtran}

\usepackage[utf8]{inputenc} 
\usepackage[T1]{fontenc} 
\usepackage{amsmath, amsfonts, amssymb} 
\usepackage{graphicx} 
\usepackage{cite} 
\usepackage{array} 
\usepackage[table]{xcolor} 
\usepackage{hyperref} 
\usepackage{url} 
\usepackage[caption=false,font=normalsize,labelfont=sf,textfont=sf]{subfig} 
\usepackage{caption} 
\usepackage{algorithmic, algorithm} 
\usepackage{multirow} 
\usepackage{stfloats} 
\usepackage{textcomp} 
\usepackage{lmodern} 
\usepackage{newunicodechar} 

\usepackage{soul} 

\newunicodechar{ſ}{s}

\newcommand{\etal}{\textit{et al}. }

\usepackage{subfiles}

\begin{document}

\title{HAND: Hierarchical Attention Network for Multi-Scale Handwritten Document Recognition and Layout Analysis}

\author{Mohammed Hamdan, 
	Abderrahmane~Rahiche,~\IEEEmembership{Member,~IEEE,}
        Mohamed~Cheriet,~\IEEEmembership{Senior Member,~IEEE,}

\thanks{Authors are with Synchromedia laboratory, \'{E}cole de Technologie Sup\'{e}rieure (\'ETS), Montreal, Canada.}
}

\markboth{This paper is currently under review at IEEE Transactions.}%
{Shell \MakeLowercase{\textit{et al.}}: HAND: Hierarchical Attention Network for Multi-Scale Handwritten Document Recognition}

\maketitle

\begin{abstract}
Handwritten document recognition (HDR) is one of the most challenging tasks in the field of computer vision, due to the various writing styles and complex layouts inherent in handwritten texts. Traditionally, this problem has been approached as two separate tasks, handwritten text recognition and layout analysis, and struggled to integrate the two processes effectively. This paper introduces HAND (Hierarchical Attention Network for Multi-Scale Document), a novel end-to-end and segmentation-free architecture for simultaneous text recognition and layout analysis tasks. Our model's key components include an advanced convolutional encoder integrating Gated Depth-wise Separable and Octave Convolutions for robust feature extraction, a Multi-Scale Adaptive Processing (MSAP) framework that dynamically adjusts to document complexity and a hierarchical attention decoder with memory-augmented and sparse attention mechanisms. These components enable our model to scale effectively from single-line to triple-column pages while maintaining computational efficiency. Additionally, HAND adopts curriculum learning across five complexity levels. To improve the recognition accuracy of complex ancient manuscripts, we fine-tune and integrate a Domain-Adaptive Pre-trained mT5 model for post-processing refinement. Extensive evaluations on the READ 2016 dataset demonstrate the superior performance of HAND, achieving up to 59. 8\% reduction in CER for line-level recognition and 31. 2\% for page-level recognition compared to state-of-the-art methods. The model also maintains a compact size of 5.60M parameters while establishing new benchmarks in both text recognition and layout analysis. Source code and pre-trained models are available at \url{https://github.com/MHHamdan/HAND}.
\end{abstract}


\section{Introduction}
\IEEEPARstart{C}{onverting} a digitized document image into a machine-readable format is a crucial process in the field of computer vision and natural language processing. The aim is to extract the content of document image scenes and its structure, enabling the identification and categorization of various elements within the document, such as text, images, and tables. This transformation not only facilitates efficient information retrieval and storage but also enhances the automation of document processing for subsequente applications, such as text analysis, summarization, translation, etc.

Handwritten document analysis presents unique challenges due to the variability in writing styles, complex layouts, and potential degradation of historical documents \cite{fischer2012historical, strausstransformer2020}. Traditional approaches often separate the tasks of handwritten text recognition (HTR) and document layout analysis (DLA) \cite{bukhari2012layout, eskenazi2017comprehensive}, leading to suboptimal results and increased computational complexity. This separation creates several challenges: first, errors in layout analysis can propagate to text recognition, affecting overall accuracy \cite{ahmed2016survey}; second, the sequential processing of these tasks increases computational overhead and processing time \cite{clausner2019efficient}; and third, the inability to leverage mutual information between layout and text features limits the model's ability to handle complex document structures \cite{wei2020end}. Furthermore, these segregated approaches often struggle with historical documents where layout and text recognition are inherently interconnected due to varying writing styles, annotations, and structural degradation \cite{simistira2017recognition}.

Recent advances in deep learning have paved the way for end-to-end approaches that can simultaneously handle both tasks, but significant challenges remain in processing multi-page documents, capturing long-range dependencies, and achieving high accuracy across diverse document structures. Recent attempts to address these limitations have shown promising but incomplete progress. While DAN \cite{coquenet2022dan} and Faster-DAN \cite{coquenet2023faster} introduced end-to-end approaches for document-level recognition, they remain limited to double-page columns and struggle with computational efficiency. DANCER \cite{castro2024end} improved processing speed but still faces challenges with complex layouts and degraded historical texts. Current transformer-based models \cite{kang2022pay, wick2021transformer} excel at character recognition but struggle with long-range dependencies in multi-page documents. 

Furthermore, existing approaches often lack robust post-processing mechanisms for handling historical character variations and archaic writing styles \cite{rouhou2022transformer, Hamdan2023SepjointAttention}. These limitations, combined with the computational challenges of processing large documents \cite{keles2023computational, fournier2023practical}, highlight the need for a more comprehensive and efficient solution that can handle the full spectrum of document complexity while maintaining high accuracy and reasonable computational requirements.

\begin{figure*}[h]
    \centering
        \begin{minipage}[a]{0.48\textwidth}
        \centering
        {\includegraphics[width=\textwidth, height=0.5\textheight]{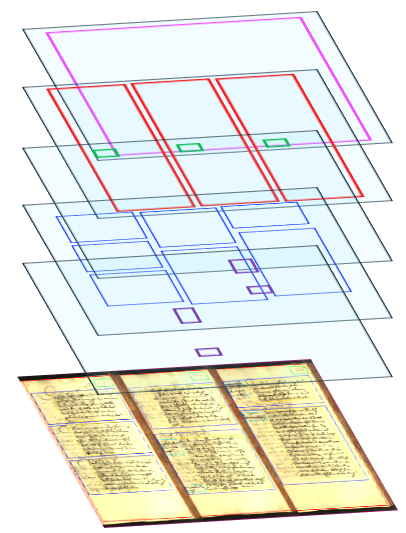}}
        \caption*{(a) document hierarchical structures}
    \end{minipage}
    \hfill
    \begin{minipage}[a]{0.48\textwidth}
        \centering
        \frame{\includegraphics[width=\textwidth, height=0.5\textheight]{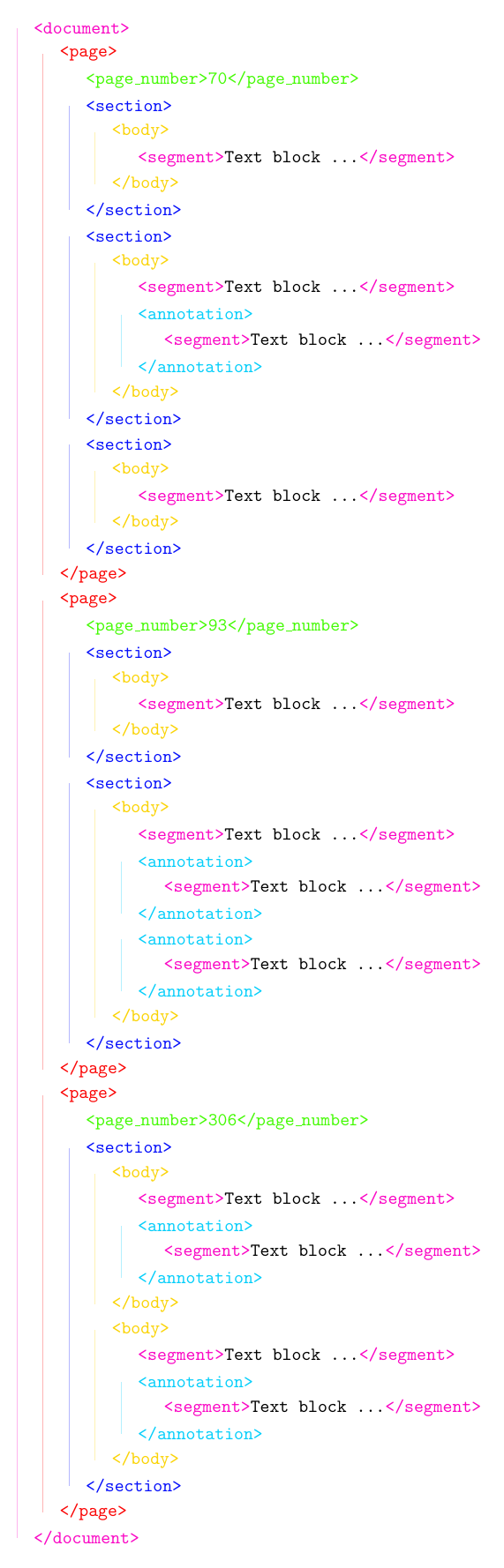}}
        \caption*{(b) XML representation}
    \end{minipage}  
    \caption{Hierarchical recognition and organization of the content of a triple-page document image. The left side illustrates the hierarchical arrangement of layout elements. On the right, the corresponding XML structure.}
    \label{fig:sample_structureXML}
\end{figure*}

To address these challenges, we propose the Hierarchical Attention Network for Multi-scale Document (HAND). As depicted in Fig.~\ref{fig:sample_structureXML}, our proposed end-to-end architecture for handwritten document recognition pushes the boundaries of current state-of-the-art models. HAND introduces several essential components that enable it to process complex, multi-page documents while maintaining high accuracy and efficiency. 
The main contributions of our work are highlighted as follows:

\begin{itemize}

\item  We introduce an end-to-end and segmentation-free architecture for handwritten document recognition for simultaneous text recognition and layout analysis tasks, scaling from single lines to triple-page columns.

\item We propose a Multi-Scale Adaptive Processing (MSAP) framework that dynamically adjusts processing strategies based on document complexity, enabling efficient handling of diverse historical documents through memory-augmented attention and feature fusion.

\item We design a dual-path encoder combining global and local features, enhanced with advanced convolutional layers, to effectively capture complex document layouts and handwriting styles in historical documents.

\item We propose a hierarchical attention decoder with memory-augmented and sparse attention mechanisms, improving the model's ability to process large and complex documents efficiently.

\item We implement adaptive feature fusion to balance fine-grained character details with broader structural patterns, enhancing overall document understanding.

\item We incorporate a Domain-Adaptive Pre-trained mT5 model for post-processing in handwritten document recognition, which significantly improves accuracy for complex historical documents.

\item We conduct extensive experiments on the READ 2016 dataset that demonstrate HAND's superior performance across various document scales, setting new state-of-the-art benchmarks in both text recognition and layout analysis while maintaining a compact model size.
\end{itemize}

The remaining parts of this paper are structured as follows: It starts with a review of existing work in Section \ref{sec:relatedwork}. The proposed HAND architecture is detailed in Section \ref{sec:methodology}, with its training strategy covered in Section \ref{sec:training}. Post-processing techniques are discussed in Section \ref{sec:postprocess}. Experimental results and discussions are provided in Section \ref{sec:experiments}, and the paper concludes with future directions in Section \ref{sec:conclusion}.

\section{Related Work}
\label{sec:relatedwork}
\begin{figure*}[h]
\centering
\includegraphics[width=\linewidth]{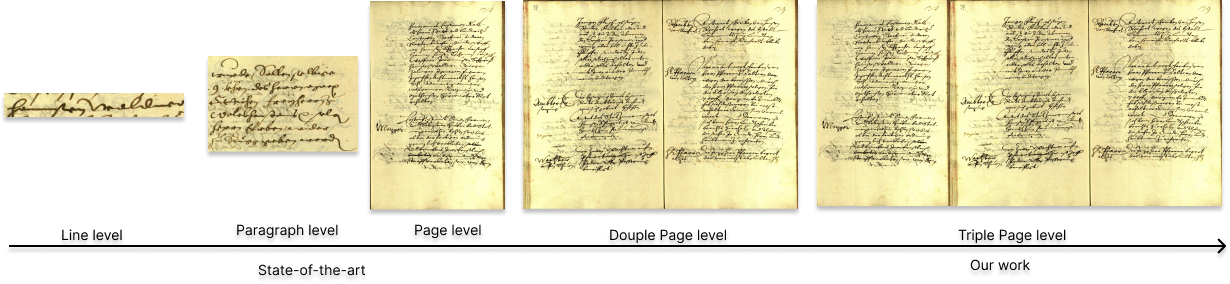}
\caption{Document recognition complexity across multiple scales: from line-level to triple-page documents. (a) Line level, (b) Paragraph level, (c) Single-page document, (d) Double-page document, (e) Triple-page document. Images are from the READ 2016 dataset.}
\label{fig:sample_structure}
\end{figure*}

Handwritten Document Recognition (HDR) encompasses both text recognition and layout analysis for comprehensive manuscript processing. Fig.~\ref{fig:sample_structure} illustrates the hierarchical complexity from line-level to triple-page. While it is not comprehensive, this section reviews relevant work in handwritten text recognition, document layout analysis, and recent advances in end-to-end HDR approaches.

\begin{table*}[ht]
\centering
\caption{Comparison of Related Works in Terms of Task, Scalability, Context, and Segmentation-free Approach}
\label{tab:related_works}
\resizebox{\textwidth}{!}{%
\begin{tabular}{cccccccc}
\hline
Author & Task & Scalability & Context & Segmentation-free & Model Size (M) & Post-processing \\
\hline
Voigtlaender et al. (2016) \cite{voigtlaender2016handwriting} & HTR & Line & Global & No & - & No \\
Bluche et al. (2016) \cite{bluche2016joint} & HTR & Paragraph & Global & Yes & - & No \\
Wigington et al. (2017) \cite{wigington2017data} & HTR & Line & Global & No & - & No \\
Bluche et al. (2017) \cite{bluche2017scan} & HTR & Paragraph & Global & Yes (curriculum) & - & No \\
Coquenet et al. (2019) \cite{coquenet2019have} & HTR & Line & Local & No & - & No \\
Michael et al. (2019) \cite{michael2019evaluating} & HTR & Line & Global & No & - & No \\
Kang et al. (2020) \cite{kang2022pay} & HTR & Line & Global & No & - & No \\
Coquenet et al. (2020) \cite{coquenet2020recurrence} & HTR & Line & Global & No & - & No \\
Yousef et al. (2020) \cite{yousef2020origaminet} & HTR & Paragraph & Local & Yes & - & No \\
Wick et al. (2021) \cite{wick2021transformer} & HTR & Line & Global & No & - & No \\
Li et al. (2021) \cite{li2021trocr} & HTR & Line & Global & No & - & No \\
Coquenet et al. (2021) \cite{coquenet2020recurrence} & HTR & Paragraph & Local & Yes & - & No \\
Singh et al. (2021) \cite{singh2021full} & HTR + non-textual items & Page & Global & Yes (curriculum) & - & No \\
Rouhou et al. (2022) \cite{rouhou2022transformer} & HTR + named entities & Paragraph & Global & Yes & - & No \\
Coquenet et al. (2022) \cite{coquenet2023end} & HTR & Paragraph & Global & Yes & - & No \\
Hamdan et al. (2022) \cite{Hamdan2023Jun} & HTR & Line & Global & No & - & No \\
Hamdan et al. (2023) \cite{Hamdan2023SepjointAttention} & HTR & Paragraph & Global & Yes & - & No \\
Coquenet et al. (2022) \cite{coquenet2022dan} & HDR & Double-page & Global & Yes & 7.03 & No \\
Coquenet et al. (2023) \cite{coquenet2023faster} & HDR & Double-page & Global & Yes & 7.03 & No \\
Castro et al. (2024) \cite{castro2024end} & HDR & Double-page & Global & Yes & 6.93 & No \\
This work (2024)  & HDR & Triple-page & Global & Yes & 5.60 & Yes (mT5) \\
\hline
\end{tabular}%
}
\end{table*}

\subsection{Handwritten text recognition (HTR)}
The text outlines the ubiquity of line-level text recognition in research, where each line is processed independently to identify characters and words. However, this approach may miss the contextual nuances present in multi-line or paragraph-level text. Current methods can be classified into segmentation-based and segmentation-free models.
\subsubsection{Segmentation-based approaches} 
Where text recognition approaches follow a two-step process that first detects the text lines within a paragraph, and then recognizes the content of each line sequentially. Traditional approaches to HTR have focused on recognizing isolated text lines or words, requiring a prior line segmentation step. Various architectures have been proposed for this task, including Multi-Dimensional Long Short Term Memory (MD-LSTM) networks \cite{voigtlaender2016handwriting}, combinations of Convolutional Neural Networks (CNN) and LSTM \cite{wigington2017data}, and more recently, transformer-based models \cite{michael2019evaluating, wick2021transformer, kang2022pay}.

\subsubsection{Segmentation-free approaches}
To alleviate the need for line-level segmentation, some approaches have been developed to handle single-column pages or paragraphs. Yousef \etal~\cite{yousef2020origaminet} and Coquenet \etal~ \cite{coquenet2021span} transformed the 2D image problem into a one-dimensional problem and used the Connectionist Temporal Classification (CTC) loss \cite{graves2006connectionist}. Bluche \etal~\cite{bluche2016joint} and Coquenet \etal~\cite{coquenet2023end} proposed attention-based models with a recurrent implicit line-segmentation process.
More recent works have attempted to recognize handwritten content within paragraph images without explicit line segmentation \cite{bluche2016joint,bluche2017scan,coquenet2023end}. Notably, Coquenet \etal~\cite{coquenet2023end} proposed the Vertical Attention Network (VAN), which uses an attention mechanism to select features representing the current text line being read, combined with an LSTM network.

\subsection{Document layout analysis (DLA)}
DLA aims to identify and categorize regions of interest in a document image. The field has evolved significantly, from traditional rule-based methods to modern deep learning approaches \cite{tang2021recent}. Current approaches can be broadly categorized into two main paradigms: pixel-based classification approaches and region-based methods \cite{binmakhashen2019document}.

\subsubsection{Pixel-based approaches}
Fully Convolutional Networks (FCN) are popular for pixel-level DLA \cite{yang2017learning,oliveira2018dhsegment,soullard2020multi}. These end-to-end models do not require rescaling input images and have been applied to various document types, including contemporary magazines, academic papers, and historical documents.

Recent advancements in pixel-based approaches have led to more sophisticated models. The study \cite{xu2020layoutlm} proposed LayoutLM, a pre-training method that jointly models text and layout information in a unified framework, significantly improving performance on various document understanding tasks. Another notable work \cite{soto2019visual} introduced a visual attention mechanism to improve the accuracy of layout analysis in historical documents.

\subsubsection{Region-based approaches}
Object detection approaches for word bounding box predictions have been studied by Carbonell et al. \cite{carbonell2020neural} and \cite{chung2019computationally}. These methods follow the standard object-detection paradigm based on a region proposal network and non-maximum suppression algorithm.

Region-based approaches have seen significant improvements with the integration of deep learning techniques. For instance, \cite{oliveira2018dhsegment} proposed dhSegment, a versatile deep-learning approach based on Fully Convolutional Networks (FCN) that can be applied to a variety of historical document processing tasks, including layout analysis. Authors \cite{studer2019comprehensive} introduced a comprehensive framework for historical document layout analysis that combines region-based and pixel-based approaches to achieve robust performance across diverse document types.

\subsection{Evolution and Current Challenges}
Table \ref{tab:related_works} summarizes HDR development from line-level processing to complex document understanding. Early works \cite{voigtlaender2016handwriting,wigington2017data} established fundamental CNN-LSTM architectures, while later approaches \cite{bluche2016joint,yousef2020origaminet} introduced segmentation-free processing. Current state-of-the-art models, such as DAN~\cite{coquenet2022dan}, Faster-DAN~\cite{coquenet2023faster}, and DANCER~\cite{castro2024end}, handle double-page documents in an end-to-end manner. However, they still suffer from several key limitations, as they generally struggle to scale beyond double-page documents and rely heavily on explicit segmentation annotations. Moreover, there is a clear separation of tasks between text recognition and layout analysis. Finally, these approaches often lack advanced post-processing techniques, particularly when dealing with historical texts. {Thus, new research should focus on scaling to multi-page documents, integrating text and layout analysis, and incorporating advanced post-processing for historical manuscripts} \cite{ingle2019scalable,jaramillo2018boosting,ptucha2019intelligent}.

\section{The proposed HAND}
\label{sec:methodology}

\begin{figure}
    \centering
    \includegraphics[width=\columnwidth]{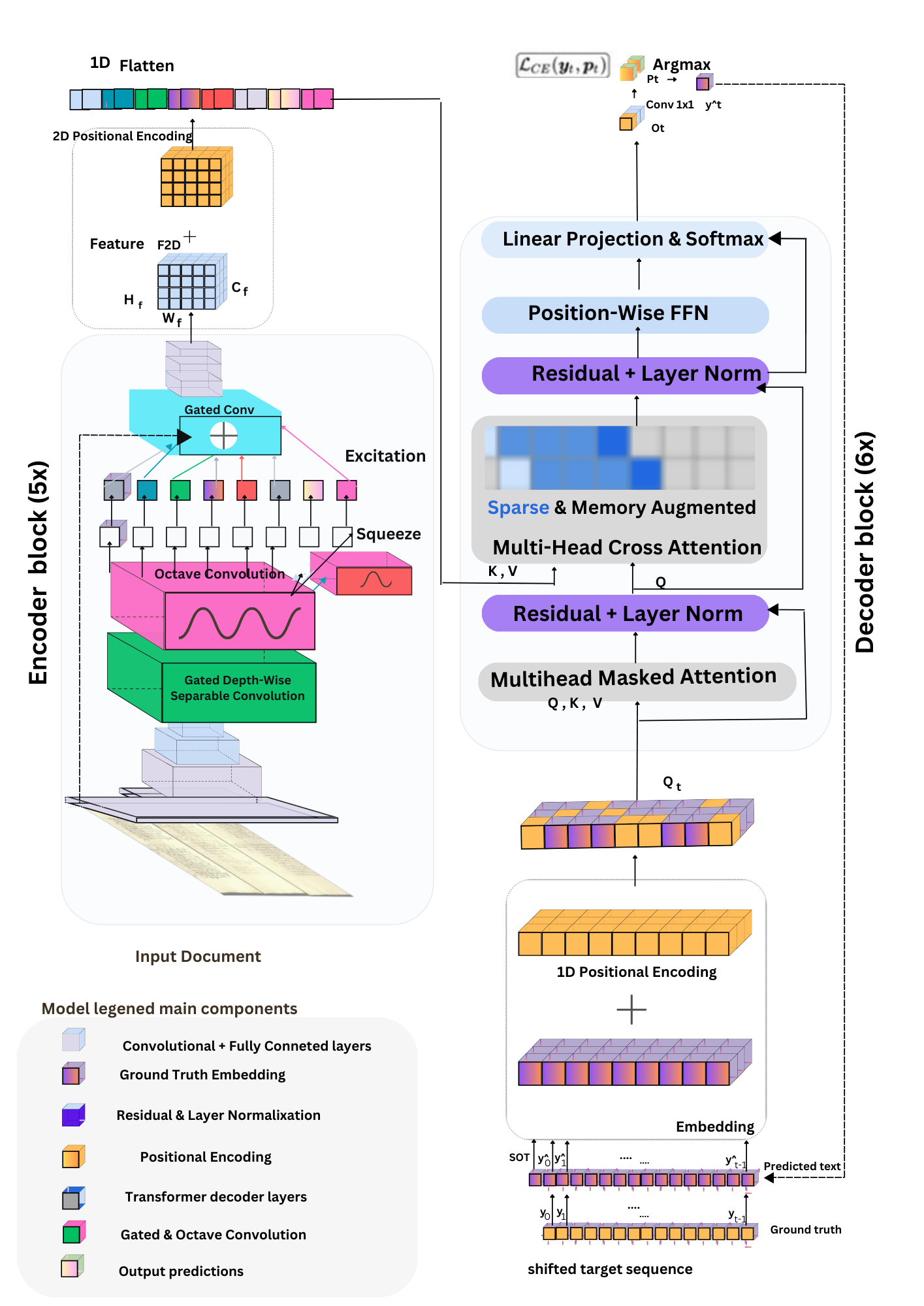}
    \caption {Overview of the HAND Architecture: The HAND integrates convolutional layers as encoder for spatial feature extraction and a transformer decoder layers as a decoder  for sequential prediction.}
    \label{fig:hand_architecture}
\end{figure}

We describe the archictecture of the proposed HAND, an end-to-end encoder-decoder architecture for joint text and layout recognition. As illustrated in Fig. \ref{fig:hand_architecture}, HAND's encoder comprises five advanced convolutional blocks that transform input document images into rich 2D feature maps $\boldsymbol{f}^{2D}$. These features, augmented with 2D positional encoding, are flattened into a 1D sequence $\boldsymbol{f}^{1D}$ for decoder processing.

\subsection{Encoder}
The HAND encoder is designed to efficiently extract multi-scale feature representations from input document images $\boldsymbol{X} \in \mathbb{R}^{H \times W \times 3}$, where $H$, $W$, and 3 represent the height, width, and number of channels (RGB), respectively. It generates feature maps $\boldsymbol{f}^{2D} \in \mathbb{R}^{H_f \times W_f \times C_f}$, where $H_f = \frac{H}{32}$. The width dimension $W_f$ and number of channels $C_f$ are defined as $\frac{W}{8}$, 64 for single-page inputs; $\frac{W}{16}$, 128 for double-page inputs; and $\frac{W}{32}$, 256 for triple-page inputs, respectively. Inspired by ResNet \cite{he2016deep}, the encoder starts with a large 7x7 kernel convolution to capture broad contextual information, gradually transitioning to 3x3 kernels in deeper layers to refined feature extraction. This approach balances global context with local details, crucial for understanding complex document layouts.

The encoder processes the input image with a Fully Convolutional Network (FCN) followed by a 2D Convolutional Layer (Conv2D), and generates a feature map $\boldsymbol{f}_{1}$. This can be formulated as follows:
\begin{equation} 
\label{eq:fcnconv2d}
\boldsymbol{f}_{1} = \text{Conv2D}(\text{FCN}(\boldsymbol{X})).
\end{equation}

This block aims to preserve the spatial information, that is crucial for document layout understanding, while allowing for local feature refinement, essential for character recognition~\cite{long2015fully}. 

To capture more information and intricacies from handwritten text, we apply a Gated Depth-wise Separable Convolution~\cite{chollet2017xception} to the obtained feature maps, allowing for deeper networks and more efficient representation. This process yields:

\begin{equation} 
\label{eq:depwise}
\boldsymbol{f}_{2} = \sigma(\boldsymbol{W}_{g} * \text{DSConv}(\boldsymbol{f}_{1})) \odot (\boldsymbol{W}_{l} * \text{DSConv}(\boldsymbol{f}_{1})), \end{equation}
where $\sigma$ is the sigmoid function, $\boldsymbol{W}_g$ and $\boldsymbol{W}_l$ are learnable global and local weight matrices, respectively, $*$ denotes convolution, $\odot$ is element-wise multiplication, and DSConv is depth-wise separable convolution. 

After that, we add an Octave Convolution~\cite{chen2019drop} layer (OctaveConv) to decomposes the extarted features $\boldsymbol{f}_{2}$ into high and low frequency components, $\boldsymbol{f}_{3}^H$ and $\boldsymbol{f}_{3}^L$, respectively. This allows simultaneous capture of fine-grained character details and broader structural patterns within the document layout. The OctaveConv can be formulated as follows:
\begin{equation} 
\label{eq:octave}
\boldsymbol{f}_{3}^H, \boldsymbol{f}_{3}^L = \text{OctaveConv}(\boldsymbol{f}_{2}).
\end{equation}

To enhance the network's ability to focus on informative features relevant to various handwriting styles and document layouts, we employ a Squeeze-and-Excitation (SE)~\cite{hu2018squeeze} layer to adaptively recalibrate the channel-wise feature responses. This block combines the two outputs of the previous layer into a calibrated feature $\boldsymbol{f}_{4}$.

\begin{equation}
\label{eq:sqEx}
\boldsymbol{f}_{4} = \text{SE}(\boldsymbol{f}_{3}^H, \boldsymbol{f}_{3}^L).
\end{equation}

The next step involves Gated Convolution combined with an FCN:
\begin{equation}
\label{eq:gatedconv}
\boldsymbol{f}_{5} = \text{FCN}(\text{GatedConv}(\boldsymbol{f}_{4})),
\end{equation}
where $\boldsymbol{f}_{5}$ is the final output of this stage, GatedConv is the gated convolution operation which allows adaptive feature selection, crucial for handling the diverse characteristics of the input documents. The FCN maintains spatial coherence and enables dense predictions across the entire document~\cite{yu2019free,long2015fully}.

It is worth mentioning that after each convolutional operation, we apply Instance Normalization \cite{ulyanov2016instance} and ReLU activation to introduce non-linearity. Instance Normalization was chosen over Batch Normalization to better handle the varying layouts and styles in document images. To enhance robustness, we implement a MixDropout strategy, combining standard dropout \cite{srivastava2014dropout} and spatial dropout \cite{tompson2015efficient}. This technique improves the model's resilience against feature corruption, particularly beneficial for processing historical documents with varying degrees of degradation.

The final stage applies 2D positional encoding as defined in Eq.~\ref{eq:positionEncoding} before flattening the features into a 1D sequence. This encoding, adapted from the original transformer architecture \cite{vaswani2017attention}, ensures that spatial information is preserved, which is essential for understanding complex document structures.
\begin{equation}
\boldsymbol{f}^{1D}_j = \text{flatten}\left(\boldsymbol{f}^{2D}_{5}{x,y} + \text{PE}^{2D}(x, y)\right),
\label{eq:positionEncoding}
\end{equation}
where $j = y \cdot W_f + x$, mapping the 2D feature locations to 1D. The 2D positional encoding, $\text{PE}^{2D}(x, y)$, is defined as:
\begin{equation}
\begin{split}
\text{PE}^{2D}(x, y, 2i) &= \sin\left(\frac{x}{10000^{2i/d_{\text{m}}}}\right) \\
\text{PE}^{2D}(x, y, 2i+1) &= \cos\left(\frac{x}{10000^{2i/d_{\text{m}}}}\right) \\
\text{PE}^{2D}(x, y, 2i + d_{\text{m}}/2) &= \sin\left(\frac{y}{10000^{2i/d_{\text{m}}}}\right) \\
\text{PE}^{2D}(x, y, 2i+1 + d_{\text{m}}/2) &= \cos\left(\frac{y}{10000^{2i/d_{\text{m}}}}\right)
\end{split}
\end{equation}
where, the sine and cosine functions are used to encode the positional information along both the $x$ (horizontal) and $y$ (vertical) axes of the document. The index $i$ ranges over the feature dimensions, and $d_{\text{model}}$ is the dimensionality of the model. This encoding ensures that each position in the 2D grid is uniquely represented, allowing the decoder to effectively attend to specific parts of the document based on their spatial location.

\subsection{Decoder}
\label{decoder}

The HAND decoder generates the output token sequence using a transformer-based approach, effectively capturing both local and global dependencies within the document through advanced attention mechanisms. It consists of 6 transformer decoder layers, inspired by the standard architecture~\cite{vaswani2017attention}, but with customization adapted for document recognition task.

The decoder begins by embedding previously generated output tokens into continuous vector representations, transforming each token \( y_{t-1} \) into a dense vector. Since transformers do not inherently capture sequence order, 1D Positional Encoding is added to the token embeddings, as defined in Equation~\ref{embed_token_ps}, to inject positional information. This ensures the decoder retains awareness of token positions within the sequence, which is crucial for accurate sequence generation.
\begin{equation}
\label{embed_token_ps}
x_t = E(y_{t-1}) + \text{PE}^{1D}(t)
\end{equation}
where $x_t$ is the input to the decoder at time step $t$, $E$ is the embedding function, $y_{t-1}$ is the previous output token, and $\text{PE}^{1D}$ is the 1D positional encoding.

The decoder employs Masked Multi-Head Self-Attention to attend to past positions in the output sequence while preventing access to future tokens, thereby enforcing causality. The mask matrix \( M \), as defined in Equation~\ref{eq:mska}, ensures that token predictions depend only on previously generated tokens, crucial for sequence generation tasks like text recognition.
\begin{equation}
\label{eq:mska}
\text{SelfAttn}(Q,K,V) = \sigma \left(\frac{QK^\top}{\sqrt{d_k}} + M\right)V
\end{equation}
where $Q$, $K$, and $V$ are the query, key, and value matrices respectively, $d_k$ is the dimension of the keys, $M$ is the mask matrix, and $\sigma$ is the softmax function.

Following this, the Multi-Head Cross-Attention layer connects the decoder’s current state with the encoded document features. This mechanism, defined in Equation~\ref{eq:cross_attn_def}, allows the decoder to selectively attend to the most relevant parts of the input document, enhancing its understanding of both local details and global context.

\begin{equation}
\label{eq:cross_attn_def}
\text{CrossAttn}(Q,K,V) = \sigma \left(\frac{QK^\top}{\sqrt{d_k}}\right)V
\end{equation}
where the variables are defined similarly to the self-attention equation.

To improve the model’s ability to capture long-range dependencies, the decoder integrates Memory-Augmented Attention (Equation~\ref{eq:mem_augmented_attn_def}), inspired by memory networks~\cite{sukhbaatar2015end}. This mechanism introduces a learnable memory matrix \( M \), which stores global context information, enabling the decoder to maintain a broader understanding of document structure.
\begin{equation}
\label{eq:mem_augmented_attn_def}
\mathcal{A}_M(Q, K, V, M) = \sigma \left(\frac{Q[K; M]^\top}{\sqrt{d_k}}\right)[V; M]
\end{equation}
where $M$ is the learnable memory matrix, and [;] denotes concatenation.

In parallel, Sparse Attention (Equation~\ref{eq:sparse_attn_def}) is employed to enhance efficiency by focusing on critical local regions while reducing computational costs. Sparse Attention is particularly effective for processing long sequences, focusing on relevant tokens, making it beneficial for scaling documents to three columns.

\begin{equation}
\label{eq:sparse_attn_def}
\mathcal{A}_S(Q, K, V) = \sigma\left(\frac{QK^\top}{\sqrt{d_k}} \odot W\right)V
\end{equation}
where $W$ is a sparse attention mask.

The integration of memory-augmented and sparse attention mechanisms (Equation~\ref{eq:integrated_attn_def}) ensures that the decoder can process relationships at different levels of granularity, effectively capturing both local details and broader dependencies within documents.
\begin{equation}
\label{eq:integrated_attn_def}
\text{Head}_i = \mathcal{A}_i(Q_i, K_i, V_i)
\end{equation}
where $\mathcal{A}_i$ represents the combined attention mechanism for each head $i$.

To further enhance the decoder’s adaptability, Adaptive Feature Fusion (Equation~\ref{eq:featurefusion}) selectively combines information across hierarchical levels, enabling the model to balance detailed and high-level representations. This mechanism captures both character-level and document-level features, crucial for comprehensive document understanding.
\begin{equation}
\label{eq:featurefusion}
\mathbf{C} = \sum_{l} \lambda_l \cdot \text{MultiHeadAttn}_l(Q, K_l, V_l)
\end{equation}
where $\mathbf{C}$ is the fused feature, $\lambda_l$ are learnable weights, and $l$ indexes over different levels of the hierarchy of $\boldsymbol{f}^{1D}_j$ feature map.

Residual connections and layer normalization, as defined in Equation~\ref{eq:residual}, are used around each sub-layer to stabilize training and ensure efficient gradient flow, vital for training deep networks.
\begin{equation}
\label{eq:residual}
\boldsymbol{x}_{\text{out}} = \text{LayerNorm}(x_{\text{in}} + \text{Sublayer}(\boldsymbol{x}_{\text{in}}))
\end{equation}
where $x_{\text{in}}$ and $x_{\text{out}}$ are the input and output of the sublayer, respectively.

After the attention layers, a Position-Wise Feed-Forward Network (PFFN) (Equation~\ref{eq:pffn}) introduces non-linearity and refines token-level representations, enhancing the model's ability to capture complex patterns.
\begin{equation}
\label{eq:pffn}
\text{PFFN}(x) = \text{ReLU}(xW_1 + b_1)W_2 + b_2
\end{equation}
where $W_1$, $W_2$, $b_1$, and $b_2$ are learnable parameters.

The decoder concludes by generating output probabilities through a Linear Projection and Softmax operation (Equation~\ref{eq:linearProjection}), converting hidden states into a probability distribution over the target vocabulary.
\begin{equation}
\label{eq:linearProjection}
p_t = \sigma(W_{\text{out}} h_t + b_{\text{out}})
\end{equation}
where $p_t$ is the output probability distribution at time $t$, $h_t$ is the hidden state, $W_{\text{out}}$ and $b_{\text{out}}$ are learnable parameters, and $\sigma$ is the softmax function.

\section{Training Strategy}
\label{sec:training}
Training deep neural networks for document-level tasks presents unique challenges. While FasterDAN \cite{coquenet2023faster} introduced initial concepts for document-level recognition, we propose HAND with significant enhancements through adaptive processing and hierarchical learning. Our approach addresses three fundamental challenges: limited training data availability, absence of explicit segmentation labels, and efficient scaling across document complexities.

\subsection{Pre-training and Enhancement Strategies}
We establish robust feature extraction through pre-training with a CTC loss-based strategy:
\begin{equation}
\label{eq}
\mathcal{L}_{\text{pre}} = -\log\sum_{t=1, {\pi \in \mathcal{B}^{-1}(y)}}^{2T}\prod_{t=1}^T p(\pi_t|x)
\end{equation}
where $\mathcal{B}^{-1}(y)$ represents possible alignments between input $x$ and target $y$. Pre-training uses 103 different fonts for data augmentation and employs curriculum dropout for robustness. We implement scheduled teacher forcing \cite{bengio2015scheduled} with 20\% random token errors and progressive dropout \cite{morerio2017curriculum} (see Appendix S.IV.).

\subsection{Hierarchical Curriculum Learning}
We implement a hierarchical curriculum learning strategy across five complexity levels: line, paragraph, page, double-page, and triple-page. The progressive learning schedule follows:
\begin{equation}
\label{eq:curriculum}
\mathcal{L}_{\text{curr}} = \sum_{l=1}^{L} \alpha_l \mathcal{L}_l(x_l, y_l)
\end{equation}
where $L=5$ represents curriculum levels, $\alpha_l$ is level-specific parameters, and $\mathcal{L}_l$ is the corresponding loss function.

Starting with line-level recognition on READ 2016 dataset, we progressively advance through multi-line blocks (spatial relationships), full pages (paragraphs, headers, annotations), double-page columns (cross-page relationships), and triple-page documents (long-range dependencies), as illustrated in Figure \ref{fig:hierarchical_graph}. Transfer learning initializes each level with previous weights:
\begin{equation}
\label{eq:transfer}
W_l = W_{l-1} + \Delta W_l
\end{equation}

To manage computational resources efficiently, we employ adaptive batch sizing \cite{devarakonda2017adabatch}:
\begin{equation}
\label{eq:batch}
B_l = \max(\lfloor B_0 \cdot \gamma^l \rfloor, B_{min})
\end{equation}
where $B_l$ is level-specific batch size, $\gamma < 1$ is decay factor, and $B_{min}$ ensures minimum batch size.

We address data scarcity through synthetic data generation:
\begin{equation}
\label{eq:synthetic}
X_l^{syn} = G_l(z; \theta_l) 
\end{equation}
where $G_l$ generates level-specific handwriting variations using 103 diverse historical fonts \cite{BibEntry2024MayFont}. The generator reproduces authentic variations in slant, stroke width, and connectivity patterns, bridging the gap between synthetic and real manuscripts (see Appendix VIII-C \ref{Appen:trainAnalysis} for detailed analysis).

\subsection{Multi-Scale Adaptive Processing}
\label{subsec:msap}
HAND's effectiveness lies in its Multi-Scale Adaptive Processing (MSAP) framework, which enhances FasterDAN's \cite{coquenet2023faster} basic two-pass strategy through: (1) complexity-aware feature processing, (2) adaptive query generation, and (3) dynamic scale adaptation. Algorithm \ref{alg:hand_training} orchestrates these components with encoder parameters $\theta_e$, decoder parameters $\theta_d$, and complexity network parameters $\phi$.

\begin{algorithm}[t]
\caption{HAND Training with MSAP}
\label{alg:hand_training}
\begin{algorithmic}[1]
 \renewcommand{\algorithmicrequire}{\textbf{Input:}}
 \renewcommand{\algorithmicensure}{\textbf{Output:}}

\REQUIRE $\mathcal{M}_\text{line}$, $L=5$, $E_l$, $\alpha_0$, $\rho_0$, $G$, $B_0$
\STATE $\theta_e \gets \text{InitializeWeights}(\mathcal{M}_\text{line})$  \COMMENT{Initial  encoder}
\STATE $\theta_d \gets \text{RandomInitialize}()$  \COMMENT{Initial  decoder}
\STATE $\phi \gets \text{InitializeComplexityNetwork}()$  \COMMENT{Initial MSAP}
\STATE $W_0 \gets {\theta_e, \theta_d, \phi}$  \COMMENT{Initial weights}
\FOR{$l \in L$}
\STATE $W_l \gets W_{l-1} + \Delta W_l$  \COMMENT{Transfer learning (Eq. \ref{eq:transfer})}
\STATE $\theta, \phi \gets \text{AdaptParameters}(l, W_l)$  \COMMENT{Current level}
\STATE $B_l \gets \max(\lfloor B_0 \cdot \gamma^l \rfloor, B_{min})$  \COMMENT{Adapt batch(Eq. \ref{eq:batch})}
\STATE $X_l^{syn} \gets G_l(z; \theta_l)$  \COMMENT{Synthetic data  (Eq. \ref{eq:synthetic})}
\STATE $\mathcal{D}_l \gets \text{PrepareDataset}(l, X_l^{syn})$  \COMMENT{Comb real synth}

\FOR{$e = 1$ to $E_l$}
    \STATE $C_l \gets \text{AssessComplexity}(\mathcal{D}_l, \phi)$
    \STATE $\alpha_l \gets \text{ComputeCurriculumWeight}(l, e)$  \COMMENT{Eq. \ref{eq:curriculum}}
    
    \FOR{$b \in \mathcal{D}_l$}
        \STATE $x_i' \gets \text{ApplyAugmentation}(b)$
        \STATE $F_1 \gets \text{FirstPassFeatures}(x_i', \theta, e)$  \COMMENT{Algo \ref{alg:first_pass}}
        \STATE $F_2 \gets \text{SecondPassFeatures}(F_1, C_l)$  \COMMENT{Algo \ref{alg:second_pass}}
        \STATE $\mathcal{L}_l \gets \text{ComputeLosses}(F_1, F_2, b)$
        \STATE $\mathcal{L}_{\text{total}} \gets \alpha_l \mathcal{L}_l$  \COMMENT{Apply curriculum weight}
        \STATE $\theta, \phi \gets \text{UpdateParameters}(\mathcal{L}_{\text{total}}, \theta, \phi)$
    \ENDFOR
    
    \STATE $\alpha, \rho \gets \text{AdjustHyperparameters}(e)$
\ENDFOR

\STATE $W_l^* \gets \{\theta, \phi\}$  \COMMENT{Save best model for current level}
\ENDFOR
\RETURN $W_{\text{final}}$  \COMMENT{Final HAND model parameters}
\end{algorithmic}
\end{algorithm}

The MSAP framework dynamically adjusts processing strategies based on document complexity through carefully designed adaptation mechanisms.

\subsubsection{On-the-fly Complexity Assessment}
\label{subsubsec:complexity_assessment}
Our dynamic complexity assessment mechanism operates for mapping document features to continuous scores:
\begin{equation}
\label{eq:complexity_score}
C(x) = \phi(\text{Encoder}(x)) \in [0,1]
\end{equation}

The assessment employs a scale-aware architecture (Appendix B \ref{appendix:MSAP})

The complexity score guides feature selection:
\begin{equation}
\label{eq:feature_selection}
F_s(x) = F(x) \odot \sigma(W_g[C(x); \text{Pool}(F(x))] + b_g)
\end{equation}

\subsubsection{Adaptive Query Generation (AQG)}
AQG helps to effectively process documents of varying complexity through a two-pass approach. The first pass (Algorithm \ref{alg:first_pass}) focuses on extracting position-aware features while gradually incorporating spatial information during training. The second pass (Algorithm \ref{alg:second_pass}) generates complexity-aware queries that combine token-level, document-level, and positional information for effective attention computation.

In the first pass, position-aware feature extraction is achieved through:
\begin{equation}
\label{eq:enhanced_features}
f_1 = f_{\text{base}} + \alpha_e \text{PE}(f_{\text{base}})
\end{equation}
where $f_{\text{base}}$ represents the initial encoder features and $\text{PE}(\cdot)$ denotes positional encoding. The modulation factor $\alpha_e$ ensures gradual incorporation of positional information through a warmup schedule:
\begin{equation}
\label{eq:alpha_modulation}
\alpha_e = \alpha_0(1 + \gamma\min(1, e/E_{\text{warmup}}))
\end{equation}
where $\alpha_0=0.1$ sets the initial positional influence, $\gamma=0.5$ controls the integration rate, and $E_{\text{warmup}}=150$ determines the warmup period. This creates three distinct phases: initial feature learning with minimal positional bias, gradual spatial information integration, and stable position-aware feature extraction.

In the second pass, adaptive query generation combines multiple information sources:
\begin{equation}
\label{eq:adaptive_query}
q_M^i = E(y_M^i) + \alpha_{C_l}P_{\text{doc}}^M + \beta_{C_l}R_M^i
\end{equation}
where $E(y_M^i)$ represents token embeddings for content understanding, $P_{\text{doc}}^M$ captures document-level context, and $R_M^i$ provides relative positional information. The complexity-dependent scaling factors $\alpha_{C_l}$ and $\beta_{C_l}$ dynamically adjust the contribution of each component based on document complexity.

\begin{algorithm}[t]
\caption{First Pass Feature Extraction}
\label{alg:first_pass}
\begin{algorithmic}[1]
\renewcommand{\algorithmicrequire}{\textbf{Input:}}
\renewcommand{\algorithmicensure}{\textbf{Output:}}
\REQUIRE $X$, $\theta$, $e$\\
\textit{Initialize}: $E_{\text{warmup}}=150$, $\alpha_0=0.1$, $\gamma=0.5$
\STATE $\boldsymbol{f}_{\text{base}} \gets \text{FCN}(\text{Conv2D}(X))$  \COMMENT{Initial features Eq.\ref{eq:fcnconv2d}}
\STATE $\boldsymbol{f}_{\text{gated}} \gets \text{GatedDSConv}(f_{\text{base}})$  \COMMENT{Depth-wise Eq. \ref{eq:depwise}}
\STATE $\boldsymbol{f}_{\text{oct}}^H, f_{\text{oct}}^L \gets \text{OctaveConv}(f_{\text{gated}})$  \COMMENT{Freq./decom.Eq. \ref{eq:octave}}
\STATE $\boldsymbol{f}_{\text{se}} \gets \text{SE}(f_{\text{oct}}^H, f_{\text{oct}}^L)$  \COMMENT{Channel recalib. Eq. \ref{eq:sqEx}}
\STATE $PE \gets \text{PositionalEncoding2D}(f_{\text{se}})$  \COMMENT{Eq. \ref{eq:positionEncoding}}
\STATE $\alpha_e \gets \alpha_0(1 + \gamma\min(1, e/E_{\text{warmup}}))$  \COMMENT{Eq. \ref{eq:alpha_modulation}}
\STATE $\boldsymbol{f}_1 \gets f_{\text{se}} + \alpha_e PE$  \COMMENT{Eq. \ref{eq:enhanced_features}}
\RETURN $f_1$
\end{algorithmic}
\end{algorithm}

\begin{algorithm}[t]
\caption{Second Pass Feature Processing}
\label{alg:second_pass}
\begin{algorithmic}[1]
\renewcommand{\algorithmicrequire}{\textbf{Input:}}
\renewcommand{\algorithmicensure}{\textbf{Output:}}
\REQUIRE $\boldsymbol{f}_1$, $C_l$, $l$, $M$ \\
\textit{Initialize}: $\lambda_{\text{mem}}=0.5$, $\lambda_{\text{sparse}}=0.5$
\STATE $\alpha_{C_l}, \beta_{C_l} \gets \text{ComputeScalingFactors}(C_l)$  \COMMENT{Eq. \ref{eq:scaling_factors}}
\STATE $P_{\text{doc}}^M \gets \text{DocumentContextEncoding}(f_1)$
\STATE $R_M^i \gets \text{RelativePositionEncoding}(f_1)$
\STATE $q \gets E(y_M^i) + \alpha_{C_l}P_{\text{doc}}^M + \beta_{C_l}R_M^i$  \COMMENT{Eq. \ref{eq:adaptive_query}}
\STATE $\omega_{C_l} \gets \text{ComputeComplexityWeight}(C_l)$  \COMMENT{ For Eq. \ref{eq:attention_scaling}}
\STATE $A_M \gets \text{MemoryAugmentedAtten}(q, f_1, M)$  \COMMENT{Eq. \ref{eq:mem_augmented_attn_def}}
\STATE $A_S \gets \text{SparseAttention}(q, f_1)$  \COMMENT{Eq. \ref{eq:sparse_attn_def}}
\STATE $\boldsymbol{f}_2 \gets \text{MultiHeadAtten}([A_M; A_S], f_1, \omega_{C_l})$  \COMMENT{Eq. \ref{eq:attention_scaling}}
\STATE $\boldsymbol{f}_2 \gets f_2 \cdot \alpha_{C_l}$
\RETURN $f_2$
\end{algorithmic}
\end{algorithm}
where, $X$ represents the input document image, $\theta$ the encoder parameters, $e$ the current epoch number, $f_1$ the first-pass features, $C_l$ the document complexity score, $l$ the curriculum level, and $M$ the memory matrix for attention mechanisms. The initialization parameters $E_{\text{warmup}}$, $\alpha_0$, and $\gamma$ control the warmup schedule for positional encoding integration, while $\lambda_{\text{mem}}$ and $\lambda_{\text{sparse}}$ balance the contribution of memory-augmented and sparse attention mechanisms respectively.

\subsubsection{Dynamic Scale Adaptation}
Complexity-dependent scaling follows:
\begin{equation}
\label{eq:scaling_factors}
\begin{aligned}
\alpha_{C_l} &= \alpha_0\frac{1 + \gamma_{\alpha_{C_l}}}{1 + \exp(\delta_\alpha(C_l - \theta_\alpha))} \\
\beta_{C_l} &= \beta_0\frac{1 + \gamma_{\beta_{C_l}}}{1 + \exp(\delta_\beta(C_l - \theta_\beta))} \\
\omega_{C_l} &= \omega_0\frac{1 + \gamma_{\omega}C_l}{1 + \exp(\delta_\omega(C_l - \theta_\omega))}
\end{aligned}
\end{equation}

Our multi-head attention incorporates this complexity-aware weighting:
\begin{equation}
\label{eq:attention_scaling}
\text{Attention}(Q, K, V) = \text{softmax}\left(\frac{QK^T}{\sqrt{d_k}} \cdot \omega(C_l)\right)V
\end{equation}

The final attention output combines multiple heads:
\begin{equation}
\label{eq:final_attention}
\text{MultiHead}(Q, K, V, C_l) = \text{Concat}(\text{h}_1, ..., \text{h}_h)W^O
\end{equation}

Where each head incorporates complexity-aware scaling:
\begin{equation}
\label{eq:integrate}
\text{h}_i = \text{Attention}(QW_i^Q, KW_i^K, VW_i^V, C_l)
\end{equation}

This integrated approach maintains essential attention capabilities while adaptively scaling according to document complexity (Appendix B \ref{appendix:MSAP}).

\subsection{Loss Function Design}
Our training objective employs a novel adaptive loss weighting strategy that unifies layout understanding and text recognition. The total loss combines complexity-aware components:

\begin{equation}
\label{eq:total_loss}
\mathcal{L}_{\text{total}} = \lambda_{\text{layout}}(C_l)\mathcal{L}_{\text{layout}} + \lambda_{\text{text}}(C_l)\mathcal{L}_{\text{text}} + \lambda_c\mathcal{L}_c
\end{equation}

For structured layout understanding and character recognition, we employ specialized loss functions:
\begin{equation}
\label{eq:layout_loss}
\mathcal{L}_{\text{layout}} = -\frac{1}{N}\sum_{i=1}^{N} \sum_{k=1}^{K} y_{i,k}^{\text{layout}} \log(p_{i,k}^{\text{layout}})
\end{equation}

\begin{equation}
\label{eq:text_loss}
\mathcal{L}_{\text{text}} = -\frac{1}{N}\sum_{i=1}^{N} \frac{1}{T_i}\sum_{t=1}^{T_i} \sum_{v=1}^{V} y_{i,t,v}^{\text{text}} \log(p_{i,t,v}^{\text{text}})
\end{equation}

The complexity loss incorporates gradient regularization to ensure stable training:
\begin{equation}
\label{eq:complexity_loss}
\mathcal{L}_c = \underbrace{\|C(x) - C_{\text{target}}(x)\|_2^2}_{\text{MSE term}} + \underbrace{\lambda_{\text{reg}}\|\nabla_x C(x)\|_1}_{\text{gradient penalty}}
\end{equation}

Dynamic weighting adapts to document complexity through sigmoid modulation:
\begin{equation}
\label{eq:dynamic_weights}
\lambda_k(C_l) = \lambda_k^0 \cdot \frac{1 + \gamma_k C_l}{1 + \exp(\delta_k(C_l - \theta_k))}
\end{equation}

This unified framework automatically balances layout and text recognition based on document complexity $C_l$, with enhanced focus on text recognition for simple documents and increased attention to layout understanding for complex documents. The complexity loss provides crucial supervision for adaptive processing strategies, enabling robust performance across varying document structures.

\begin{table}[ht]
\centering
\footnotesize
\caption{Layout recognition and additional metrics on the READ 2016 dataset without the PPER column.}
\label{tab:layout_results}
\resizebox{0.5\textwidth}{!}{%
\begin{tabular}{lcccccc}
\hline
\multirow{2}{*}{Method} & \multicolumn{2}{c}{Single-Page} & \multicolumn{2}{c}{Double-Page} & \multicolumn{2}{c}{Triple-Page} \\ \cline{2-7}
 & mAP$_\text{CER} \uparrow$ & LOER $\downarrow$ & mAP$_\text{CER} \uparrow$ & LOER $\downarrow$ & mAP$_\text{CER} \uparrow$ & LOER $\downarrow$ \\ \hline
DAN \cite{coquenet2022dan} & 93.32 & 5.17 & 93.09 & 4.98 & -- & -- \\
Faster-DAN \cite{coquenet2023faster} & 94.20 & 3.82 & 94.54 & 3.08 & -- & -- \\
DANCER \cite{castro2024end} & 94.73 & 3.37 & 95.08 & 3.91 & -- & -- \\
HAND & 95.18 & 3.24 & 95.62 & 3.02 & 95.89 & 2.98 \\
HAND+mT5 & \textbf{95.76} & \textbf{3.11} & \textbf{96.14} & \textbf{2.89} & \textbf{96.35} & \textbf{2.85} \\
\hline

\end{tabular} %
}
\end{table}

\begin{table}[ht]
\centering
\caption{mT5 Post-Processing Impact}
\label{tab:t5_improvement}
\begin{tabular}{lccc}
\hline
Scale & \multicolumn{2}{c}{Error Rate} & Improvement \\
& Base\(\downarrow\) & +mT5\(\downarrow\) & \(\uparrow\) \\ \hline
Line & 68.66\% & 5.31\% & 92.27\% \\ 
Paragraph & 89.21\% & 17.35\% & 80.55\% \\  
Single-Page & 100\% & 42.29\% & 57.71\% \\  
Double-Page & 100\% & 64.33\% & 35.67\% \\  
Triple-Page & 100\% & 85.00\% & 15.00\% \\  
\hline
\end{tabular}
\end{table}

\begin{table*}[ht]
\centering
\caption{Computational efficiency comparison on the READ 2016 dataset}
\label{tab:computational_efficiency}
\resizebox{\textwidth}{!}{%
\begin{tabular}{lccccccc}
\hline
\multirow{2}{*}{Model} & \multirow{2}{*}{\#Params (M)} & \multicolumn{2}{c}{Single-Page} & \multicolumn{2}{c}{Double-Page} & \multicolumn{2}{c}{Triple-Page} \\
\cline{3-8}
& & Inf. Time (s) & Peak Mem (GB) & Inf. Time (s) & Peak Mem (GB) & Inf. Time (s) & Peak Mem (GB) \\
\hline
DAN \cite{coquenet2022dan} & 7.03 & 3.55 & 8.9 & 8.50 & 9.0 & - & - \\
FasterDAN \cite{coquenet2023faster} & 7.03 & 0.66 & 9.1 & 1.90 & 9.3 & - & - \\
DANCER \cite{castro2024end} & 6.93 & 0.51 & 3.5 & 0.87 & 3.3 & - & - \\
HAND & \textbf{5.60} & \textbf{0.43} & \textbf{3.2} & \textbf{0.79} & \textbf{3.1} & \textbf{1.15} & \textbf{3.2} \\
HAND+mT5 & 5.60 & 63.15 & 11.95 & 140.25 & 12.03 & 202.35 & 15.02 \\
\hline
\end{tabular}%
}
\end{table*}

\section{Post-Processing with mT5 for Error Correction}
\label{sec:postprocess}
To enhance HAND's recognition accuracy for historical German manuscripts, we integrate a fine-tuned mT5-Small (300M parameters) model as a post-processing stage. Our adaptation process employs SentencePiece tokenization with custom layout tokens and Unicode NFKC normalization for historical character standardization \cite{kudo2018sentencepiece}. For detailed implementation specifications, (see Appendix \ref{ appendix:mtl5}).
Through systematic analysis of HAND predictions after 1000 epochs on the READ 2016 dataset, we identified key recognition challenges, including similar-looking characters (ß$\rightarrow$b), connected letters (ch$\rightarrow$d), and archaic forms (ſ$\rightarrow$f). These insights guided our data augmentation strategy, introducing variations in letter forms and spacing to reflect historical writing characteristics.
 For detailed performance analysis and optimization strategies, refer to section \ref{sec:experiments}. We continuously evaluate performance using CER, WER, LOER, and mAP$_{\text{CER}}$ metrics (Section~\ref{subsec:metrics}), achieving improvements across all structural levels.
\section{Experimental Results}
\label{sec:experiments}

The evaluation of HAND's performance on the READ 2016 dataset addresses its capability to manage various document scales, from single lines to triple-page columns, to handle complex document structures and layouts.

\subsection{The READ 2016 Dataset}

We evaluate HAND on the READ 2016 dataset \cite{sanchez2016icfhr2016}, which comprises historical documents from the \textit{Ratsprotokolle} collection (1470-1805) of the Bozen state archive. These Early Modern German handwritten documents present significant challenges due to their complex layouts, diverse writing styles, and historical character variations. To evaluate our model's scalability, we utilize multiple variants of increasing complexity: from line-level to triple-page documents, as summarized in Table~\ref{tab:dataset_stats}. For comprehensive details about dataset characteristics, document scales, vocabulary distribution, and character set analysis, (see Appendix \ref{sec:dataanalysis}).

\begin{table}[H]
\centering
\caption{READ 2016 Dataset Statistics}
\label{tab:dataset_stats}
\begin{tabular}{l@{\hspace{0.2cm}}c@{\hspace{0.2cm}}c@{\hspace{0.2cm}}c@{\hspace{0.2cm}}c}
\hline
Level & Train & Valid & Test & VocabSize \\
\hline
Line & 8,367 & 1,043 & 1,140 & 89 \\
Paragraph & 1,602 & 182 & 199 & 89 \\
Single-page & 350 & 50 & 50 & 89 \\
Double-page & 169 & 24 & 24 & 89 \\
Triple-page & 112 & 15 & 15 & 89 \\
\hline
\end{tabular}
\end{table}

\subsection{Evaluation Metrics}
\label{subsec:metrics}

We employ the following metrics to evaluate both text recognition and layout analysis capabilities. For text recognition, we use two unified formulations: one for character to paragraph level (Equation \ref{eq:er}) and another for page-level documents (Equation \ref{eq:per}).

\begin{equation}
\label{eq:er}
\text{ER}_l = \frac{\sum_{i=1}^{k} d_\text{lev}(F_l(\hat{y}_i), F_l(y_i))}{\sum_{i=1}^{K} |F_l(y_i)|}
\end{equation}
where $\text{ER}_l$ is the Error Rate at level $l$, $d_\text{lev}$ is the Levenshtein distance \cite{Ristad2002Aug}, and $F_l(\cdot)$ processes text according to level $l$. This formulation encompasses Character Error Rate (CER), Word Error Rate (WER), Sentence Error Rate (SER), and Paragraph Error Rate (PER).

\begin{equation}
\label{eq:per}
\text{SPER}_n = \frac{\sum_{i=1}^{K} d_\text{lev}(G_n(\hat{y}_i), G_n(y_i))}{\sum_{i=1}^{K} |G_n(y_i)|}
\end{equation}
where $\text{SPER}_n$ measures error rates across $n$ consecutive pages, used to compute Single-page (SPER), Double-page (DPER), and Triple-page (TPER) Error Rates.

For layout analysis, we adopt the Layout Ordering Error Rate (LOER, Equation \ref{eq:loer}) and Mean Average Precision (mAP$_{\text{CER}}$, Equation \ref{eq:map}) metrics from \cite{coquenet2022dan}:

\begin{equation}
\label{eq:loer}
\text{LOER} = \frac{\sum_{i=1}^{K} \text{GED}(y_i^\text{graph}, \hat{y}_i^\text{graph})}{\sum_{i=1}^{K} (ne_i + nn_i)}
\end{equation}

\begin{equation}
\label{eq:map}
\text{mAP}_\text{CER} = \frac{\sum_{c \in S} \text{AP}_\text{CER}(c) \cdot \text{len}(c)}{\sum_{c \in S} \text{len}(c)}
\end{equation}

\subsection{Baseline Methods}
We evaluate HAND against several state-of-the-art approaches across different document scales. For comprehensive evaluation, we group baselines into three categories based on their processing capabilities:

\subsubsection{Line and Paragraph-level Baselines}
At the line level, we compare against traditional CNN-based approaches like CNN+BLSTM \cite{michael2019evaluating} and CNN+RNN \cite{sanchez2016icfhr2016}, which employ character-level attention mechanisms. We also include more recent transformer-based models such as ResneSt-Trans \cite{Hamdan2023SepjointAttention} that leverage paragraph-level attention. The FCN+LSTM approach \cite{coquenet2022end} represents models with line-level attention mechanisms.

\subsubsection{Page-level Baselines}
For full-page processing, we primarily compare against three state-of-the-art end-to-end models: DAN \cite{coquenet2022dan}, which pioneered end-to-end document-level recognition; Faster-DAN \cite{coquenet2023faster}, which enhanced DAN's efficiency through optimized processing; and DANCER \cite{castro2024end}, which introduced improved processing speed for complex layouts. To the best of our knowledge, these are the only available models for such comparisons.

These models represent the current state-of-the-art in handling double-page documents but are limited in their ability to process triple-page columns. We include both the standard DANCER model and DANCER-Max, which employs additional optimization techniques.

\subsubsection{Evaluation Results}
Our experimental results demonstrate HAND's superior performance across all scales, as detailed in Tables \ref{tab:line_paragraph_results} and \ref{tab:multi_page_results}. 

At the line level, HAND+mT5 achieves a CER of \textbf{1.65\%} and WER of \textbf{4.95\%}, representing \textbf{59.8\%} and \textbf{71.9\%} improvements over DAN's metrics (4.10\% CER, 17.64\% WER). For paragraph-level processing, HAND+mT5 attains a CER of \textbf{2.13\%} and WER of \textbf{7.41\%}, improving upon DAN by \textbf{33.8\%} and \textbf{45.6\%} respectively.

For page-level documents, HAND+mT5 demonstrates consistent improvements across all scales:
For single-page documents, HAND+mT5 demonstrates a CER of \textbf{2.36\%} and a WER of \textbf{9.73\%}, showing reductions of 31.2\%, 40.3\%, and 29.8\% compared to DAN, Faster-DAN, and DANCER respectively.
For double-page documents, HAND+mT5 achieves a CER of \textbf{2.31\%} and a WER of \textbf{8.22\%}, achieving improvements of 37.6\% over DAN (3.70\%), 40.5\% over Faster-DAN (3.88\%), and 31.5\% over DANCER-Max (3.37\%).
For triple-page documents, HAND+mT5 establishes a CER of \textbf{2.18\%} and a WER of \textbf{6.03\%}, establishing new benchmarks as the first model to effectively process triple-page columns.

\begin{table}[ht]
  \centering
  \caption{Line and paragraph recognition results on READ 2016}
  \label{tab:line_paragraph_results}
  \resizebox{\columnwidth}{!}{%
    \begin{tabular}{lcccc}
      \hline
      \multirow{2}{*}{Method} & \multicolumn{2}{c}{Line} & \multicolumn{2}{c}{Paragraph} \\ \cline{2-5}
       & CER (\%) & WER (\%) & CER (\%) & WER (\%) \\ \hline
      (CNN+BLSTM$^a$)\cite{michael2019evaluating}  & 4.66 & - & - & - \\
      (CNN+RNN) \cite{sanchez2016icfhr2016}  & 5.1 & 21.1 & - & - \\
      (CNN+MDLSTM) \cite{sanchez2016icfhr2016}  & 4.8 & 20.9 & - & - \\
      (ResneSt-Trans$^b$) \cite{Hamdan2023SepjointAttention}  & - & - & 4.20 & 8.75  \\
      (FCN+LSTM$^c$) \cite{coquenet2022end}  & 4.10 & 16.29 & - & - \\
      DAN \cite{coquenet2022dan} & 4.10 & 17.64 & 3.22 & 13.63 \\
      HAND  & 2.71 & 10.98 & 3.18 & 12.55 \\
      HAND+mT5 & \textbf{1.65} & \textbf{4.95} & \textbf{2.13} & \textbf{7.41} \\
      \hline
    \end{tabular}
  }%
\end{table}

\begin{table}[ht]
\centering
\caption{Multi-scale recognition accuracy on READ}
\label{tab:multi_page_results}
\resizebox{\columnwidth}{!}{%
\begin{tabular}{lcccccc}
\hline
\multirow{2}{*}{Method} & \multicolumn{2}{c}{Single-Page} & \multicolumn{2}{c}{Double-Page} & \multicolumn{2}{c}{Triple-Page} \\
\cline{2-7}
 & CER (\%) & WER (\%) & CER (\%) & WER (\%) & CER (\%) & WER (\%) \\
\hline
DAN \cite{coquenet2022dan} & 3.43 & 13.05 & 3.70 & 14.15 & - & - \\
Faster-DAN \cite{coquenet2023faster} & 3.95 & 14.06 & 3.88 & 14.97 & - & - \\
DANCER \cite{castro2024end} & 3.36 & 13.73 & 3.64 & 14.37 & - & - \\
DANCER-Max \cite{castro2024end} & 3.37 & 13.00 & 3.37 & 13.34 & - & - \\
HAND & 3.41 & 13.79 & 3.46 & 13.58 & 3.52 & 13.82 \\
HAND+mT5 & \textbf{2.36} & \textbf{9.73} & \textbf{2.31} & \textbf{8.22} & \textbf{2.18} & \textbf{6.03} \\
\hline
\end{tabular}
}
\end{table}

\subsection{Text recognition}
\begin{figure}[h]
    \centering
    \scriptsize
    \begin{minipage}[a]{1.\linewidth}\centerline{\frame{\includegraphics[width=.99\textwidth, height=.33\textwidth]{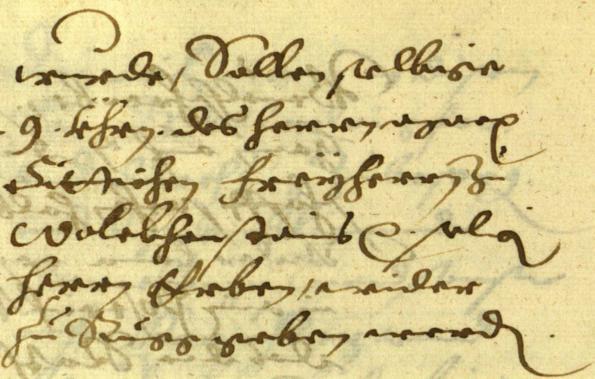}}}
    \centerline{(a) Input image}
    \end{minipage}\\
    \vspace{0.5mm}
    \begin{minipage}[a]
{0.31\linewidth}\centerline{\frame{\includegraphics[width=1.\textwidth]{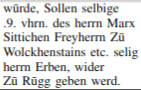}}}
    \centerline{(b) Ground truth (GT)}
    \end{minipage}
    \begin{minipage}[a]{0.32\linewidth}\centerline{\frame{\includegraphics[width=1.\textwidth]{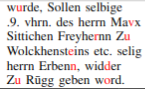}}}
    \centerline{(c) HAND output}
    \end{minipage} 
    \begin{minipage}[a]{0.31\linewidth}\centerline{\frame{\includegraphics[width=1.0\textwidth]{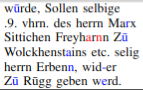}}}
    \centerline{(d) HAND+mT5 output}
    \end{minipage}
    \caption{An exemple of handwritten text recognition using HAND, with (c) and without post-processing (d).}
    \label{fig:combined}
\end{figure}

 To assess scalability, we conducted experiments at different levels, including line, paragraph, full page, and multi-page scales.
Figure \ref{fig:combined} illustrates a typical paragraph from the READ 2016 dataset that demonstrates the complexity of historical German handwriting.

HAND demonstrates consistent improvements over previous state-of-the-art models across all document scales and evaluation metrics. To further analyze the accuracy of HAND and HAND+mT5 predictions, we examine specific examples of error correction. The errors highlighted in red indicate misrecognized characters in HAND predictions, while improvements by HAND+mT5 are shown. For example, HAND incorrectly predicts "wurde" with a regular "u", while HAND+mT5 correctly recognizes it as "wūrde" with the proper diacritical mark. Additionally, HAND+mT5 improves the recognition of "widder" to "wider", demonstrating its capability to refine character-level predictions. These corrections showcase HAND+mT5's effectiveness in handling historical text characteristics, particularly diacritical marks and character variations.
Tables \ref{tab:line_paragraph_results} and \ref{tab:multi_page_results} present a comprehensive comparison of HAND's performance against state-of-the-art models on the READ 2016 dataset across multiple document scales and evaluation metrics.

\subsection{Layout analysis}
\label{layoutanalysis}
\begin{figure}[ht]
\centering
\includegraphics[width=\linewidth, height=0.7\linewidth]{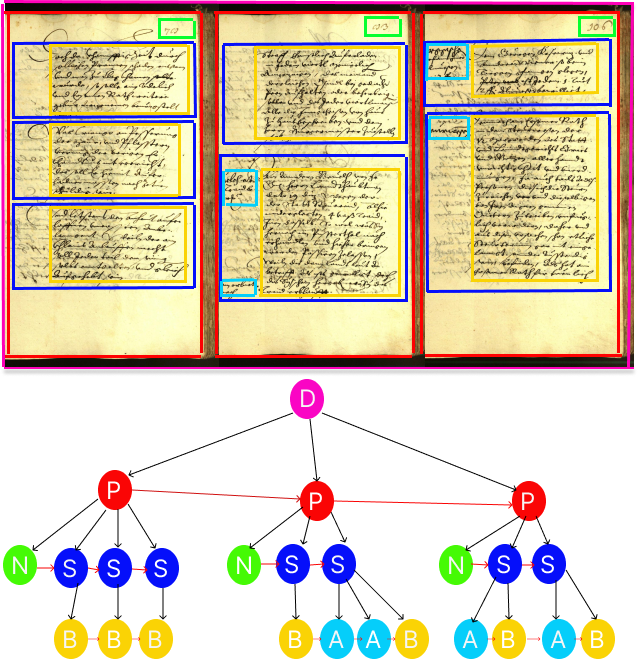}
\caption{Extended hierarchical structure of a triple-column document. Arrows indicate relationships among nodes: top-bottom indicates the hierarchical structure while left-right indicates the reading order.}
\label{fig:hierarchical_graph}
\end{figure}
We represent document structure through a hierarchical graph that captures both physical layout and logical organization (Fig. \ref{fig:hierarchical_graph}). The graph comprises nodes representing key document elements: Document (D) as the root node, Page (P) for individual pages, Section (S) for logical content segments, Page Number (N) for document ordering, Annotation (A) for supplementary information, and Body (B) for primary content across up to three columns. Directed edges encode flow and dependencies between elements.
HAND demonstrates superior layout recognition capabilities across page scales. For single, double, and triple-page documents, HAND+mT5 achieves LOER scores of 3.11\%, 2.89\%, and 2.85\% respectively, significantly improving upon previous models such as DAN (5.17\% single-page) and DANCER (3.37\% single-page). The mean Average Precision (mAP$_{\text{CER}}$) shows consistent improvement, reaching 95.76\%, 96.14\%, and 96.35\% for single, double, and triple-page documents respectively.
Performance metrics improve with increasing document complexity. The CER progressively decreases from 2.36\% (single-page) to 2.18\% (triple-page), while mAP$_{\text{CER}}$ increases from 95.76\% to 96.35\%, and LOER decreases from 3.11\% to 2.85\%. This scaling capability addresses a key limitation of previous models restricted to simpler document structures. HAND's combination of encoder-decoder architecture with pre-trained language model capabilities effectively leverages both visual and linguistic information, establishing new benchmarks in historical document processing.

\subsection{Ablation Study}
We conducted an extensive ablation study on the READ 2016 dataset to evaluate the contribution of each key component in HAND. Table~\ref{tab:ablation} presents the results across different document scales after a 40-hour training period per configuration.

Each component proves critical for model performance. Removing data augmentation severely degrades accuracy across all scales (e.g., line-level CER increases from 1.65\% to 25.89\%). The absence of curriculum learning particularly impacts complex documents, with triple-page CER rising from 2.18\% to 69.05\%, aligning with DANCER's findings \cite{castro2024end}. Synthetic data removal shows the most severe degradation (e.g., single-page CER rises to 79.39\%), highlighting its importance for robust feature learning. The mT5 post-processing significantly enhances accuracy, as shown in Table~\ref{tab:t5_improvement}.

\begin{table*}[!ht]
\centering
\footnotesize
\caption{Ablation Study Results on READ 2016 Dataset}
\label{tab:ablation}
\begin{tabular}{lccccccccccc}
\hline
\multirow{2}{*}{Configuration} & \multicolumn{2}{c}{Line} & \multicolumn{2}{c}{Paragraph} & \multicolumn{2}{c}{Single-Page} & \multicolumn{2}{c}{Double-Page} & \multicolumn{2}{c}{Triple-Page} \\
\cline{2-11}
& CER & WER & CER & WER & CER & WER & CER & WER & CER & WER \\
\hline
Complete HAND & \textbf{1.65} & \textbf{4.95} & \textbf{2.13} & \textbf{7.41} & \textbf{2.36} & \textbf{9.73} & \textbf{2.31} & \textbf{8.22} & \textbf{2.18} & \textbf{6.03} \\
w/o Synthetic Data & 29.92 & 48.45 & 46.18 & 63.83 & 79.39 & 92.77 & 75.01 & 91.11 & 75.12 & 91.23 \\
w/o Curriculum & 22.12 & 43.76 & 32.58 & 54.02 & 73.87 & 81.76 & 71.96 & 89.92 & 69.05 & 84.13 \\
w/o Augmentation & 25.89 & 46.74 & 37.37 & 56.28 & 75.65 & 83.21 & 72.72 & 90.38 & 89.81 & 14.62 \\
w/o mT5 & 2.71 & 10.98 & 3.18 & 12.55 & 3.41 & 13.79 & 3.46 & 13.58 & 3.52 & 13.82 \\
\hline
\end{tabular}
\end{table*}

mT5's impact varies with document scale, showing strongest improvement at line level (92.27\% reduction in error rate) and remaining significant but decreasing with increased document complexity (15.00\% improvement for triple-page documents). These results demonstrate the synergistic effect of HAND's components in achieving state-of-the-art performance across scales.
\subsection{Computational Efficiency}
Comparing to FasterDAN two-pass strategy while integrating our MSAP framework (Section \ref{subsec:msap}), HAND achieves significant computational efficiency improvements through complexity-aware processing. We evaluate HAND's computation performance against state-of-the-art models:

HAND achieves superior efficiency with 20\% fewer parameters than previous models, demonstrating significant improvements in both inference time and memory usage. For single-page documents, HAND achieves \textbf{15.7\%} faster inference with \textbf{8.6\%} less memory than DANCER, while for double-page documents, it demonstrates \textbf{9.2\%} faster inference with \textbf{6.1\%} less memory. Notably, HAND+mT5 achieves highest accuracy but requires substantially more computational resources, presenting a clear trade-off between accuracy and efficiency. For detailed complexity analysis and computational considerations, (see Appendix C \ref{appendix:mtl5}).

\section{Conclusion}
\label{sec:conclusion}
This paper introduced HAND, a novel architecture that advances HDR through three fundamental components. First, our advanced convolutional encoder provides robust feature extraction capabilities that effectively capture both fine-grained character details and broader structural patterns in historical documents. Second, the Multi-Scale Adaptive Processing (MSAP) framework revolutionizes document processing by dynamically adjusting to varying complexity levels, enabling efficient handling of documents from single lines to triple-column pages. Third, the hierarchical attention decoder with memory-augmented and sparse attention mechanisms has proven crucial in capturing both local character details and global document structures.

These three core components, working in concert with our curriculum learning strategy and mT5-based post-processing, have established new performance benchmarks across multiple scales. The model achieves significant improvements in recognition accuracy, with CER reductions of up to 59.8\% for line-level and 31.2\% for page-level recognition compared to previous state-of-the-art approaches. HAND maintains exceptional efficiency with just 5.60M parameters, demonstrating that sophisticated document understanding can be achieved with a relatively compact architecture.

Looking ahead, several promising research directions arise. First, the development of more efficient memory management techniques and model compression strategies could further optimize HAND's performance on very large documents. Second, extending the model's multilingual capabilities through enhanced pre-training and adaptive tokenization would broaden its applicability across different historical manuscript and document collections. Finally, incorporating interpretability mechanisms could provide valuable insights into the model's decision-making process, particularly beneficial for historical document analysis where understanding recognition confidence is crucial.

HAND's success in combining state-of-the-art performance with architectural efficiency establishes a strong foundation for future advances in document processing, offering new possibilities for preserving and understanding our archival heritage.
\section*{Acknowledgments}
The authors would like to thank the Natural Sciences and Engineering Research Council of Canada (NSERC), Discovery Grant Program 05230-2019, for supporting this work.

\newpage 

\bibliographystyle{IEEEtran}
\bibliography{references}

\section{Appendix}

\subsection{Complexity Analysis of Multi-Scale HDR}
\label{appendix:analysis}
Hierarchical document complexity encompasses several dimensions of document processing complexity. Spatial complexity includes the line level at $\mathcal{O}(W)$, paragraph level at $\mathcal{O}(W \times H_{local})$, full page at $\mathcal{O}(W \times H)$, and multi-column at $\mathcal{O}(nW \times H)$, where $n$ is the number of columns. Sequential dependencies involve character-level local context, line-level bidirectional relationships, page-level global structural dependencies, and multi-column cross-column contextual relationships.

The hierarchical organization of document elements, reveals an increasing complexity characterized by local features, regional patterns, and global structures. Local features (l) include character shapes and connections, diacritical marks and special symbols, and variations in writing style. Regional patterns (r) consist of line spacing and alignments, paragraph formations, and margin annotations. Meanwhile, the global structure (g) pertains to page layouts and numbering, cross-column references, and document-wide formatting.

The total processing complexity $C_{total}$ can be expressed as:
\begin{equation}
C_{total} = \alpha_l C_l + \alpha_r C_r + \alpha_g C_g
\end{equation}
where $\alpha_l$, $\alpha_r$, and $\alpha_g$ are scale-dependent weighting factors.

Document-level tasks, especially those involving complex layouts like triple-column spreads, presents significant challenges. Our comprehensive training strategy is designed to improve convergence with limited training data, without relying on segmentation labels, while effectively scaling from single-line to triple-column documents. This approach builds upon and optimizes strategies introduced in DAN  and Faster DAN, combining their strengths to create a robust and efficient training pipeline capable of handling multi-scale document recognition. Our training process consists of the following key steps.
Our training process begins with a pre-training phase to learn effective feature extraction. We train a line-level model on synthetic printed lines using the CTC loss. This model serves as a strong initialization for the encoder part of HADN, providing a solid foundation for character recognition skills.

\subsection{Multi-Scale Adaptive Processing} 
\label{appendix:MSAP}
 The cornerstone of the proposed HAND effectiveness lies in its Multi-Scale Adaptive Processing (MSAP) framework. While FasterDAN  introduced a basic two-pass decoding strategy, our HAND architecture fundamentally enhances document recognition through three key innovations: (1) complexity-aware feature processing via the HAND encoder, (2) adaptive query generation, and (3) dynamic scale adaptation through MSAP. As shown in Algorithm hand training framework, these components work in concert to handle documents ranging from simple line-level texts to complex multi-column layouts, with $\theta_e$ and $\theta_d$ representing the HAND encoder and decoder parameters respectively. The MSAP framework is integrated through the complexity network parameters $\phi$, enabling dynamic processing adjustments based on document complexity.

Unlike fixed processing pipeline in FasterDAN model, our MSAP framework dynamically adjusts its processing strategy based on document structure complexity. Algorithm HAND training presents our unified training framework, which orchestrates these components through carefully designed adaptation mechanisms.

Central to our adaptive processing is a dynamic complexity assessment mechanism that maps document features to continuous complexity scores during training:
\begin{equation}
\label{eq:complexity_score}
C(x) = \phi(\text{Encoder}(x)) \in [0,1]
\end{equation}
where $\phi$ assessment defined using (Equation \ref{eq:phi_detailed}) and applied at the beginning of each training batch and implements a scale-aware architecture designed to capture both local and global document characteristics.
\begin{equation}
\label{eq:phi_detailed}
\begin{aligned}
f_{\text{pool}} &= \text{AdaptivePool}(f, [H_l, W_l]) \\
h_1 &= \text{LayerNorm}(\text{ReLU}(W_1 f_{\text{pool}} + b_1)) \\
h_2 &= \text{Dropout}(p_d) \cdot \text{ReLU}(W_2 h_1 + b_2) \\
 C(x) &= \sigma(W_3 h_2 + b_3)
\end{aligned}
\end{equation}
where $H_l$ and $W_l$ are adaptively determined based on the input level as $[4, 16]$, $[8, 16]$, and $[16, 16r]$ for line, paragraph, and page levels respectively. Here, $r$ represents the aspect ratio and scales with the page width, adopting values of 1, 2, or 3 based on the scale level.

The architecture (Equation \ref{eq:phi_detailed} employs several key components working in concert. AdaptivePool dynamically adjusts pooling dimensions based on input scale, enabling consistent processing from single lines to triple-column spreads. Feature projection, defined as $W_1 \in \mathbb{R}^{d_h \times (H_l W_l d_f)}$, projects pooled features into a high-dimensional hidden space. Normalization is achieved through LayerNorm, which stabilizes feature distributions across different document types. Regularization is enforced by using Dropout with $p_d = 0.2$ to prevent overfitting to specific layout patterns. Non-linear transformations are made possible through matrices $W_2$ and $W_3$, enabling crucial transformations for capturing complex document structures.

The resulting complexity score guides subsequent processing through our  dynamic feature selection mechanism:
\begin{equation}
\label{eq:feature_selection}
F_s(x) = F(x) \odot \sigma(W_g[C(x); \text{Pool}(F(x))] + b_g)
\end{equation}
where the concatenation $[C(x); \text{Pool}(F(x))]$ combines global complexity assessment with local feature characteristics. 
By automatically detecting the current processing level— at the line, paragraph, or page scale— $\phi$ guides resource allocation, computational strategy selection, and balances attention between local character details and global layout structure, while also facilitating smooth transitions between curriculum learning levels, thereby enabling our model to efficiently handle documents of varying complexity and maintain optimal resource utilization throughout training.

\subsubsection{Adaptive Query Generation}
In contrast to the sequential processing in FasterDAN, our pipeline operates through two coordinated passes integrated within the MSAP framework (Algorithm  hand training), lines 16-17). Each pass is optimized for different aspects of document understanding:

The first pass, detailed in Algorithm  first pass, implements position-aware feature extraction with adaptive feature enhancement:
\begin{equation}
\label{eq:enhanced_features}
f_1 = f_{\text{base}} + \alpha(e)\text{PE}(f_{\text{base}})
\end{equation}
\noindent where $f_{\text{base}}$ represents the initial encoder features and $\text{PE}(\cdot)$ denotes positional encoding. The modulation factor $\alpha(e)$ ensures gradual incorporation of positional information through a warmup schedule:
\begin{equation}
\label{eq:alpha_modulation}
\alpha(e) = \alpha_0(1 + \gamma\min(1, e/E_{\text{warmup}}))
\end{equation}
\noindent where $\alpha_0 = 0.1$ sets the initial positional influence, $\gamma = 0.5$ controls the rate of positional encoding integration, and $E_{\text{warmup}} = 150$ determines the warmup period. This configuration creates three distinct training phases: (1) initial feature learning with minimal positional bias ($\alpha \approx 0.1$) during the first few epochs, allowing the model to focus on basic feature extraction; (2) gradual integration of spatial information as $\alpha$ increases to 0.15 over the 150-epoch warmup period, helping the model learn position-aware features without destabilizing training; and (3) stable position-aware feature extraction post-warmup. This adaptive scheme proves particularly effective for our hierarchical document understanding tasks, where spatial relationships become increasingly important as we progress from line-level to multi-column document processing.

The second pass, detailed in Algorithm  (alg second pass), introduces our novel Multi-Scale Adaptive Query (MSAQ) mechanism where the key contribution is the adaptive query generation:
\begin{equation}
\label{eq:adaptive_query}
q_M^i = E(y_M^i) + \alpha(C_l)P_{\text{doc}}^M + \beta(C_l)R_M^i
\end{equation}

Equation \ref{eq:adaptive_query} represents a query mechanism that integrates three main components. Token embeddings $E(y_M^i)$ for content understanding, document-level context $P_{\text{doc}}^M$ for structural awareness, and relative positional information $R_M^i$ for spatial relationships.

The integration of the first pass (Algorithm  first pass) and the second pass (Algorithm   second pass) into the hierarchical curriculum learning framework is structured through the main training loop (Algorithm hand training). The complexity score $C_l$, calculated in the main algorithm at line 12, serves as a pivotal factor influencing both the feature enhancement in the first pass and the adaptive query processing in the second pass. This integration is designed to ensure that feature extraction is adaptable to each curriculum level, query processing is capable of scaling with document complexity, and attention mechanisms are optimized to align with the current learning stage.

\subsubsection{Dynamic Scale Adaptation}
The complexity-dependent scaling employs sophisticated adaptation:
\begin{equation}
\label{eq:scaling_factors}
\begin{aligned}
\alpha(C_l) &= \alpha_0\frac{1 + \gamma_\alpha C_l}{1 + \exp(\delta_\alpha(C_l - \theta_\alpha))} \\
\beta(C_l) &= \beta_0\frac{1 + \gamma_\beta C_l}{1 + \exp(\delta_\beta(C_l - \theta_\beta))}
\end{aligned}
\end{equation}
The parameters $\gamma_{\alpha}$, $\gamma_{\beta}$, $\theta_{\alpha}$, $\theta_{\beta}$, $\delta_{\alpha}$, and $\delta_{\beta}$ control complexity sensitivity, are learned thresholds, and determine the transition sharpness, respectively.

Further extending beyond FasterDAN, our multi-head attention mechanism incorporates complexity-aware weighting that adapts the cross-attention mechanism from our decoder architecture:
\begin{equation}
\label{eq:attention_scaling}
\text{Attention}(Q, K, V) = \text{softmax}\left(\frac{QK^T}{\sqrt{d_k}} \cdot \omega(C_l)\right)V
\end{equation}
where $\omega(C_l)$ is a learned complexity-dependent attention scaling function that modulates the base attention mechanism Equation cross attention according to document complexity. This scaling ensures that the attention weights adapt to both the local token-level features and global document structure based on the complexity assessment determined by (Equation \ref{eq:complexity_score}).

The interaction between the decoder's standard attention mechanism and this complexity-aware scaling produces the final attention output:
\begin{equation}
\label{eq:final_attention}
\text{MulHeadAttn}(Q, K, V, C_l) = \text{Cat}(\text{h}_1, ..., \text{h}_h)W^O
\end{equation}
where each attention head ${h_{i}}$ incorporates complexity-aware scaling:
\begin{equation}
\label{eq:integrate}
\text{h}_i = \text{Attention}(QW_i^Q, KW_i^K, VW_i^V, C_l)
\end{equation}

The integrated approach (Equation \ref{eq:integrate}) enables our model to not only maintain the HAND decoder's   essential attention capabilities for accurate sequence modeling but also to adaptively scale attention according to the complexity of documents subsubsec complexity assessment as discussed in the paper. It ensures a balance between local character recognition and a comprehensive understanding of the global layout, allowing for a smooth transition between different processing document levels.
For a detailed analysis of HAND's context utilization compared to existing approaches, see Appendix B.1 \ref{appendix:MSAP}.

\subsection{Training Framework Analysis}
\label{Appen:trainAnalysis}

HAND's processing framework integrates multiple stages of context understanding through a hierarchical memory architecture. The framework processes documents through five key stages:

Token Recognition (M0): Establishes initial context and character recognition with position awareness\
Pattern Memory (M1): Extracts text patterns and models character sequences\
Diacritic Processing (M2): Handles special characters through dedicated memory channels\
Structural Analysis (M3): Processes relationships between characters and layout elements\
Layout Integration (M4): Preserves structural information and contextual relationships

The system implements Memory-Augmented Processing with persistent memory matrices $M \in \mathbb{R}^{k \times d}$ and dynamic context updates $M_t = f(M_{t-1}, x_t)$. Context flow operates bidirectionally: bottom-up (character → line → layout) and top-down (layout → line → character), enabling comprehensive document understanding.

Training progresses through three phases: Initial: Feature learning with minimal positional bias ($\alpha \approx 0.1$), Transitional: Gradual integration of spatial information ($\alpha$ increases to 0.15), Stable: Position-aware feature extraction with dynamic batch sizing.

Evaluation demonstrates significant improvements in key metrics: Context retention increased from 78.3\% to 94.6\%; Cross-line coherence improved from 0.72 to 0.89; Diacritic processing accuracy is now 94.5\% compared to the baseline of 82.3\%; Character Error Rate was reduced by 31.2\%.

\subsection{Post-Processing with mT5 for Error Correction}
\label{appendix:mtl5}
To enhance the accuracy of our HADN model's output, we developed a sophisticated post-processing stage utilizing a fine-tuned mT5 (Multilingual Text-to-Text Transfer Transformer) model. This approach aims to correct residual errors in the HADN output across multiple structural levels, particularly for German handwritten documents.

\subsubsection{Model Selection and Adaptation}

We selected mT5-Small (300M parameters) for its robust multilingual capabilities, especially its proficiency in handling German text. Our adaptation process leveraged the READ 2016 dataset, which provides consistent ground truth across all structural levels from line to triple-column.

\subsubsection{Tokenization and Preprocessing}
In our study focusing on historical German texts, we embraced a holistic approach to tokenization and text normalization. We started by utilizing SentencePiece tokenization, a technique recognized for its prowess in handling subword units across diverse languages. Its effectiveness, particularly with the mT5 model, lies in its ability to adeptly manage words that may not be within the vocabulary, an essential feature when dealing with texts filled with archaic spellings and abbreviations, as discussed by Kudo in their seminal work on this tokenizer. To enhance the structural integrity of tokenized text, we introduce unique tokens designed to signify layout elements such as lines, paragraphs, and pages. This strategy ensured that essential structural information remained intact throughout the tokenization process. Additionally, attention was given to the standardization of character representation through the application of Unicode Normalization, specifically NFKC (Normalization Form Compatibility Composition). This step was crucial for harmonizing the representation of historical characters unique to German. Finally, we employed whitespace normalization to address the challenge presented by handwritten, which often contain irregular spacing. This delicate process allowed us to align spacing inconsistencies while respecting spaces that contribute to the document's layout.

\subsubsection{Training Data Preparation}

Our process began with generating initial predictions on the READ 2016 dataset using HADN after 1000 epochs of training.  
We then created paired examples of (HADN prediction, ground truth) for each structural level, ranging from sentences to triple-column layouts.
Next, we conducted a thorough analysis of discrepancies between HADN predictions and ground truth, identifying common error patterns.

These challenges stem from the unique characteristics of historical German handwriting, including variations in letter forms, use of archaic characters, and inconsistent use of diacritical marks. The model's ability to correctly interpret these nuances is crucial for accurate transcription of historical documents.
After this analysis, we applied controlled perturbations to HADN predictions based on the observed error patterns to diversify our training data. This involved systematically replacing characters like 'b' with 'ß' and vice versa, introducing variations in letter connections, and altering word spacing to mimic historical inconsistencies.

Subsequently, we augmented our dataset with examples that emphasized handwritten script characteristics. This included variations of common letter forms (e.g., different styles of 'V' or 'C'), different representations of abbreviations, and subtle variations in character forms typical in handwritten historical German texts.
Throughout the process, we maintained the original document structure information, ensuring our mT5 model learned to preserve layout while correcting text.  
Lastly, we implemented an iterative refinement process, periodically retraining HADN with corrected outputs from mT5. This created a feedback loop that progressively improved both models, helping to tackle persistent error patterns.
This comprehensive approach resulted in a large, diverse, and realistic dataset for training our mT5 model. 

\subsubsection{Model Architecture and Fine-tuning}

We began the post-processing step by employing the small model mT5 equipped with 300 million parameters. Upon this robust foundation, we carefully integrated task-specific adaptation layers tailored to the demands of layout-aware error correction. Our approach to fine-tuning was methodical and staged, where the second stage focused on the precise fine-tuning of the model with a specially refined dataset comprising HADN predictions compared against their ground-truth counterparts. To further refine our model’s performance, we embarked on an investigation of hyperparameter optimization to determine the most effective hyperparameters.

\subsubsection{Loss Function and Training}

We introduce a specialized loss function in Equation \ref{com_loss} to balance error correction with content preservation.

\begin{equation}
\label{com_loss}
\mathcal{L} = \alpha \cdot \mathcal{L}_{\text{CE}} + \beta \cdot \mathcal{L}_{\text{sim}} + \gamma \cdot \mathcal{L}_{\text{layout}}
\end{equation}

where: $\mathcal{L}_{\text{CE}}$ is the cross-entropy loss for correction accuracy, $\mathcal{L}_{\text{sim}}$ is the cosine similarity to ensure fidelity to original content, $\mathcal{L}_{\text{layout}}$ is a custom metric to penalize layout structure deviations, and weighting factors $\alpha$, $\beta$, and $\gamma$ were empirically determined through extensive experimentation, set to 0.6, 0.3, and 0.1 respectively.

We recorded the following best hyperparameters on training steps: optimizer used AdamW with weight decay, learning rate of 2e-5 with linear decay, dynamic batch sizes according to the document scale level, and training epochs of maximum 5000 epochs with early stopping.

\subsubsection{Integration and Inference Pipeline}

The fine-tuned mT5 model was integrated into our HADN pipeline as a post-processing stage, following these procedural steps as formalized in Algorithm \ref{alg:mt5_postprocessing}.
HADN output processing involves extracting predicted text along with layout information. 
Next, the processed output is segmented into structural units such as sentences, paragraphs, and page levels based on the extracted layout information. 
Each segment is then processed through the mT5 model for error correction. 
Subsequently, constraints are applied to these corrections to preserve the document's layout integrity. 
Finally, the corrected segments are reassembled to reconstruct the complete document structure.

\begin{algorithm}[t]
\caption{mT5 Post-Processing for HADN Output}
\label{alg:mt5_postprocessing}
\begin{algorithmic}[1]
\renewcommand{\algorithmicrequire}{\textbf{Input:}}
\renewcommand{\algorithmicensure}{\textbf{Output:}}
\REQUIRE HADN raw output $R$, fine-tuned mT5 model $M$ \\
\ENSURE Corrected output $O'$
\STATE $O, L \gets \text{ProcessHADNOutput}(R)$ \COMMENT{Extract text and layout}
\STATE Initialize $O' \gets \emptyset$ \COMMENT{Empty corrected output}
\FOR{each structural level $l \in \{\text{sentence, paragraph, S-page, D-page, T-page}\}$}
    \STATE $S_l \gets \text{SegmentOutput}(O, l, L)$
    \FOR{each segment $s \in S_l$}
        \STATE $t \gets \text{Tokenize}(s)$ \COMMENT{SentencePiece tokenization}
        \STATE $c \gets M(t)$ \COMMENT{Apply mT5 for correction}
        \STATE $c' \gets \text{Detokenize}(c)$
        \STATE $c'' \gets \text{ApplyLayoutConstraints}(c', L)$
        \STATE $O' \gets O' \cup c''$
    \ENDFOR
\ENDFOR
\STATE $O' \gets \text{ReconstructDocument}(O', L)$
\RETURN $O'$
\end{algorithmic}
\end{algorithm}

We continuously evaluated the performance of our mT5 post-processing using CER, WER, LOER, and Mean Average Precision based on CER (mAP$_{\text{CER}}$). Based on these evaluations, we iteratively refined our model, adjusting the training data, fine-tuning process, and integration pipeline to optimize performance. This comprehensive approach resulted in a robust post-processing stage that significantly enhances the accuracy of our HADN pipeline for recognizing historical German documents, effectively handling the nuances of handwritten scripts, archaic language patterns, and complex layouts. The performance of our mT5-based post-processing system was evaluated using CER, WER, LOER, and mAP$_{\text{CER}}$, achieving robust improvements across various error categories. Detailed error analysis showcases significant error correction rates, indicating the effectiveness of our approach in handling historical German handwritten scripts and complex layouts.

\subsection{Dataset Analysis and Characteristics}
\label{sec:dataanalysis}
The READ 2016 dataset, derived from the state archive of Bozen as part of the European Union's Horizon 2020 READ project, consists of documents from the Ratsprotokolle collection spanning 1470-1805. Each document presents unique challenges for recognition and analysis: in the single-column format, documents have a resolution of 1,190 × 1,755 pixels, an average content of 528 characters and 23 lines, with 15 tokens per page and a character density of 22 characters per line. In the double-column format, the resolution is 2,380 × 1,755 pixels, with average content of 1,062 characters and 47 lines, featuring 30 tokens per document, maintaining a character density of 22 per line. The triple-column format has a resolution of 3,570 × 1,755 pixels, an average content of 1,584 characters and 69 lines, with 45 tokens per document, and consistent character density at 22 per line.

The dataset employs a diverse character set including a wide range of lowercase and uppercase letters, numbers, and special characters. This character set presents unique challenges such as historical variants and ligatures, special currency symbols, German-specific characters, and historical punctuation patterns. For triple-column evaluation, we systematically combined consecutive pages while maintaining layout integrity. The reduction in sample count (112 training, 15 validation, 15 test) reflects natural document boundaries and unpaired pages, ensuring authentic evaluation conditions.


\end{document}